\documentclass{ceurart}

\usepackage{makecell}
\usepackage{multirow}
\usepackage{subcaption}
\usepackage{longtable}
\usepackage{color, colortbl}
\usepackage{wrapfig}

\definecolor{Gray}{gray}{0.9}
\definecolor{lightGray}{gray}{0.9}

\begin{document}

\copyrightyear{2023}
\copyrightclause{Copyright for this paper by its authors.
  Use permitted under Creative Commons License Attribution 4.0
  International (CC BY 4.0).}

\conference{CHR 2023: Computational Humanities Research Conference, December 6
  -- 8, 2023, Paris, France}

\title{T5 meets Tybalt: Author Attribution in Early Modern English Drama Using Large Language Models}

\author[1]{Rebecca M. M. Hicke}[%
orcid=0009-0006-2074-8376,
email=rmh327@cornell.edu,
url=https://rmatouschekh.github.io,
]
\cormark[1]
\address[1]{Department of Computer Science, Cornell University, USA}

\author[2]{David Mimno}[%
orcid=0000-0001-7510-9404,
email=mimno@cornell.edu,
url=https://mimno.infosci.cornell.edu,
]
\address[2]{Department of Information Science, Cornell University, USA}

\cortext[1]{Corresponding author.}

\begin{abstract}
Large language models have shown breakthrough potential in many NLP domains.
Here we consider their use for stylometry, specifically authorship identification in Early Modern English drama.
We find both promising and concerning results; LLMs are able to accurately predict the author of surprisingly short passages but are also prone to confidently misattribute texts to specific authors.
A fine-tuned \texttt{t5-large} model outperforms all tested baselines, including logistic regression, SVM with a linear kernel, and cosine delta, at attributing small passages.
However, we see indications that the presence of certain authors in the model's pre-training data affects predictive results in ways that are difficult to assess.
\end{abstract}


\maketitle

\section{Introduction}

Stylometry is a key tool for computational humanities research. Author identification provides a clear test case for methods that seek to identify ``style,'' which in turn can be used to answer many questions of interest to humanists.
However, current attribution methods require substantial amounts of known-author text for training as well as large amounts of text for identification.
Large language models (LLMs) are powerful and now widely used. They develop a statistical model of language through training on a large, unorganized corpus. By encoding information from large amounts of contextual data in their parameters, they are often able to extract subtle, complex patterns from relatively short text segments. LLMs have proven useful for tasks such as detecting scenes in German dime novels \citep{zehe2021detecting}, predicting TEI/XML annotations for plain-text editions of plays \citep{Pagel2021ab}, and understanding ancient Korean documents \citep{yoo-etal-2022-hue}.

In this work we consider whether LLMs can be applied to authorship identification and whether they might allow us to stretch the boundaries of stylometry to increasingly short passages. To evaluate these questions, we consider a deliberately difficult setting: Early Modern English drama. The language of Early Modern drama is sufficiently far from contemporary English that it may be challenging for LLMs primarily trained on modern text to parse. Additionally, the culture of co-authorship and collaboration among writers during the Early Modern era often makes it difficult to distinguish stylistic delineations between individuals. 

Despite its challenges, the attribution of Early Modern drama is a well-studied field, and techniques like cosine delta \citep{smith2011improving} achieve high accuracy at identifying plays. Yet, these methods still struggle to attribute short passages of text. We are specifically interested in determining whether fine-tuned LLMs can improve performance in this area. To this end, we provide the LLM with 5 to 450 word speaker utterances for both fine-tuning and testing. The average length of utterances in our test dataset is only 28.2 words. 

We have three primary findings.
First, for short texts the fine-tuned LLM outperforms all tested baselines, including logistic regression, a support vector machine (SVM) with a linear kernel, and cosine delta. 
Accuracy varies by author and is not fully explained by the number of plays by the author in the fine-tuning set.
Second, LLMs are more prone than cosine delta to confidently misattribute texts to specific authors.
These ``scapegoat'' authors often have large vocabularies and word use similar to the corpus average.
Third, trained LLMs may be able to quantify ``style''. 
When we apply the model trained on Early Modern drama to ``attribute'' excerpts of plays written between the 1500s and 1900s, we see an increasing proportion attributed to Shakespeare, possibly suggesting a quantification of his lasting influence.

\section{Related Work}

Many different methods have been used to perform authorship attribution tasks with Early Modern drama. These include function word adjacency networks \citep{eisen2018}, multi-view learning \citep{duque2019}, clustering algorithms \citep{arefin2014, rosso2009, naeni2016novel}, and SVMs with rolling attribution \citep{plechavc2021relative}.
All of these studies attempt to attribute complete plays except \citep{plechavc2021relative}, which attributes scenes with more than 100 lines.
We are not aware of any use of large language models for Early Modern attribution.

Attempts to attribute shorter passages in Early Modern drama have been controversial. These studies include the attribution of 63 words from \textit{Macbeth} \citep{taylor2014empirical} and samples of 173 words from \textit{Henry VI, Part 1} \citep{taylor2015imitation}.
They have been critiqued \citep{freebury2020unsound, rizvi2019problem} in part because the sections of text studied were so short.
While we attempt short text attribution, we select samples broadly from many plays rather than focusing on specific passages.

Work has also been done on the attribution of short texts in different fields. Cosine similarity is effective at attributing 500 word excerpts from blogs \citep{koppel2011authorship}.
Similarly, topic models are able to attribute email and blog snippets with average length 39 and 57 words \citep{zhang2018authorship} and the Source Code Authorship Profile (SCAP) method attributes tweets of 140 characters or shorter with high accuracy \citep{layton2010authorship, azarbonyad2015time}.
None of these studies use LLMs, and all use modern datasets.

Some researchers have begun testing the feasibility of using LLMs for attribution.
These studies used the embedding output of LLMs to train custom attribution models using LSTMs \citep{huertas-tato2022} or CNNs \citep{najafi2022}.
Our work uses a simpler LLM method, in which we fine-tune the original model to directly generate author names, without the need for any additional coding or customization.
In addition, we use a corpus with less clear delineation.

\section{Data \& Methods}

We use a collection of Early Modern English drama---plays written in the 1500s and 1600s---gathered from two sources: the Folger Digital Anthology of Early Modern English Drama (EMED) \citep{emed} and the Shakespeare His Contemporaries corpus (SHC) \citep{shc}. We first gathered 367 plays from the EMED corpus and then added the 181 remaining plays from SHC.\footnote{Because the original Shakespeare His Contemporaries corpus is no longer publicly available, we have drawn these sources from a port of the original Github linked in the citation.} In order to remove features that may distinguish files from different corpora, we stripped all non-accent non-ASCII characters from the play texts and replaced them with standardized alternatives where appropriate. Each XML file offered regularized spellings of non-standard words in the play. In creating our corpus, we used the regularized spellings from the EMED corpus that agreed with the greatest number of other sources when possible and the SHC regularizations otherwise. We chose to use the regularized text for two reasons. First, we did not want the model to be able to distinguish between authors based on spelling choices. Although differences in spelling may help the model identify authors, they are not indicative of the kinds of stylistic difference we are interested in studying. Second, we hypothesized that standardizing the play texts would make them appear more similar to modern text and thus improve the model's ability to accurately tokenize the input. Finally, we removed all line breaks from the texts as the different corpora do not consistently mark them.

We then split each play into speaker utterances to create a challenging but coherent identification problem. We separated any utterance longer than 450 words into multiple samples by splitting directly after every 450th word, regardless of sentence or line breaks. We then removed utterances with fewer than 5 words. Because authors sometimes develop distinctive speaker voices within a play, we hypothesize that separating the texts by speaker utterance adds an extra layer of difficulty to the attribution task.

\begin{table*}[t]
\caption{Example of input-output pair used during fine-tuning. We include the author label as \textit{masked} text to be generated.}
\label{tab:example}
\centering
\begin{tabular}{p{0.45\linewidth} | p{0.45\linewidth}}
\hline
\textbf{Input} & \textbf{Output}\\
\hline
AUTHOR: <extra\_id\_0> | All the damnable degrees Of drinkings have you, you staggered through one Citizen. Is Lord of two fair Manors, called you master Only for Caviar. & AUTHOR: John Webster | All the damnable degrees Of drinkings have you, you staggered through one Citizen. Is Lord of two fair Manors, called you master Only for Caviar.  \\
\hline
\end{tabular}
\end{table*} 

We further reduced the training and testing corpora to maximize validity and statistical reliability. We removed all plays with fewer than 300 remaining utterances, plays by multiple authors, and plays by authors with fewer than three works in the corpus. Plays that were mislabeled as by a single author, but were actually of disputed (co-)authorship were placed into a separate subcorpus. We were thus left with 253 plays by 23 authors in the primary corpus and 23 plays in the subcorpus. Further details about the corpora are listed in the appendix.

We used these corpora to assess the capability of several different authorship attribution methods to label short texts. Specifically, we tested logistic regression, SVMs with a linear kernal, cosine delta \citep{smith2011improving}, Pythia \citep{biderman2023pythia}, Falcon \citep{penedo2023refinedweb}, and several fine-tuned T5 models \citep{raffel2019} of varying sizes. T5 is a generative large language model and the pre-trained T5 models are optimized with a masked language modeling objective. Thus, in order to fine-tune T5 to perform authorship attribution, we created a series of input and output pairs where the inputs are formatted as an utterance with the author's name masked and the corresponding outputs are the same utterances with the author's name revealed (i.e. Table \ref{tab:example}). The tag \texttt{<extra\_id\_0>} was used to mask the author's name because it follows the format of tags used during T5's pre-training regime. 
Initial experimentation found that using this tag provided good accuracy.
It is important to note that the model could emit any string, but in practice the fine-tuned model only generated author names present in our corpus except during a later application to a comparative dataset (Section \ref{sec:style-over-time}).

\begin{wrapfigure}{r}{0.7\textwidth}
    \centering
    \includegraphics[scale=0.35]{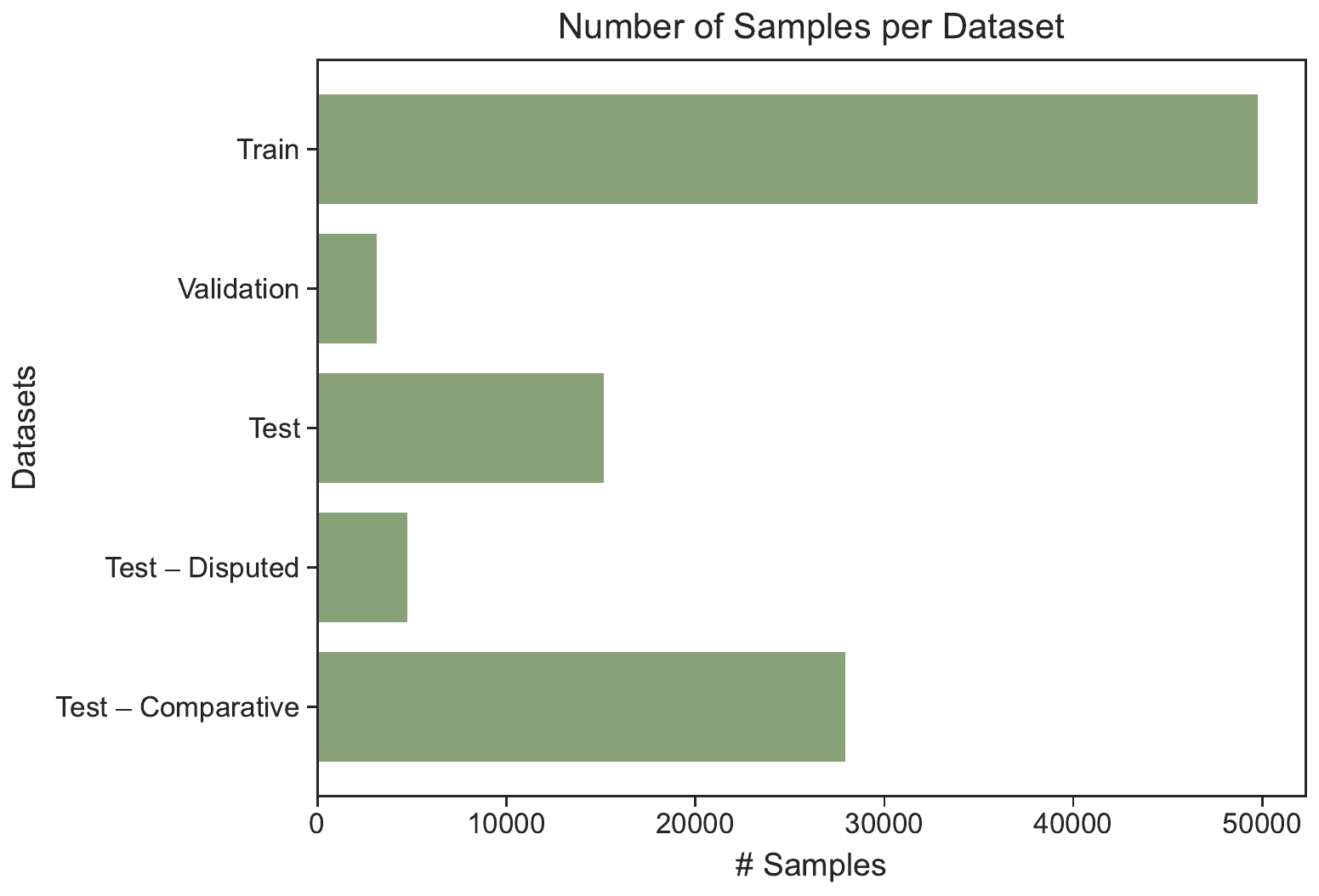}
    \caption{The size of each dataset used in the experiment by number of samples.}
    \label{fig:dataSizes}
\end{wrapfigure}

One play from each author in our corpus was withheld from the training dataset. From the remaining $n-1$ plays by each author, we included 235 random samples in the training dataset and 15 samples in the validation dataset used for parameter tuning. We included another 50 samples from each of these plays in the final test dataset for which we report results. Thus, we draw 300 distinct samples from each play withheld from the training dataset. To the test dataset, we added 200 randomly selected samples from each of the plays withheld from training. We then created a separate test dataset containing 200 samples from each of the 23 plays in the disputed authorship corpus. Figure \ref{fig:dataSizes} shows the relative size of the training, validation, and test datasets, as well as the held-out disputed set and a set of post-Early Modern plays used in Section \ref{sec:style-over-time}.

We fine-tuned a small, base, and large version of T5 on the train and validation datasets, using batch sizes of 16, 8, and 4 respectively and running for 10 epochs. The additional fine-tuning hyperparameters are reported in Section A of the appendix.
We then asked each model to predict labels for every sample in the primary test dataset. Finally, we used the best performing model, \texttt{t5-large}, to predict labels for the test dataset of disputed authorship plays. 
We also experimented with fine-tuning two comparable decoder-only generative LLMs: Pythia with 1 billion parameters and Falcon with 1 billion parameters. The input and output strings described above were edited for these experiments so that the \texttt{AUTHOR} tag was placed at the end of each string. However, both models hallucinated extensively; Pythia produced 10,221 unique strings as author names and Falcon produced 9,180. Even when the first two words of each produced string, stripped of punctuation, were used as the predicted author name Pythia and Falcon still performed considerably worse than \texttt{t5-large}. We thus omit a further analysis of these models from the paper.

For our baseline comparisons we used only the original quotation without the \texttt{AUTHOR} prefix and T5 tags. The correct authors were included as labels.
We ran two logistic regression models and two SVM models with linear kernels: one version of each used TF-IDF weighted word counts as features and the other used plain word counts. Each of these baselines was implemented using the \texttt{sklearn} package.
Cosine delta \citep{smith2011improving} is a popular improvement on Burrows delta \citep{burrows2002delta} that represents texts using z-score weighted word frequencies for the $n$ most frequent words and compares sample texts to the training corpus using cosine similarity. To run cosine delta, we used an adapted version of the \texttt{faststylometry} package with a vocabulary size of 5,000 unigrams \citep{faststylometry}. We chose a vocabulary size of 5,000 because we found it optimized performance on the plays in the training data without over-fitting and decreasing performance on the withheld plays. Each sample was assigned to the author with the highest cosine similarity value. All baseline models were evaluated on the same test/train splits as the T5 models and the TF-IDF, z-score, and word count values were fit on only the training dataset.  For every model, we experimented with using combinations of unigrams, bigrams, and trigrams but found that using only unigrams resulted in the highest performance.

\section{Comparing Models}

Results are shown in Table \ref{tab:methodCorr} for the per-sample accuracy of each attribution method and the accuracy of the ``majority vote'' predicted author of each play.
In order to display the effect of play-specific language such as character names and settings, we show predictive results for both held-out \textit{sections} of plays and fully held-out plays.

\begin{table}
\caption{LLMs have the highest predictive accuracy for short texts. The $\pm$ values represent 95\% confidence intervals. For context, we show both a lower bound (random guessing based on author frequency) and an upper bound (cosine delta on large text segments, all other rows are evaluated on short texts).} 
\small
\centering
\begin{tabular}{lccc}
\hline
\textbf{Method} & \textbf{\% Correct (In)} & \textbf{\% Correct (Out)} & \textbf{\% Correct by Play}\\
\hline
  \multicolumn{4}{c}{Upper Bound} \\
  \hline
  \rowcolor{Gray}
  \Gape[0pt][2pt]{\makecell[l]{Cosine Delta\\(Long Texts)}} & 95.8 & 87.0 & 94.9 \\
  \hline
  \multicolumn{4}{c}{LLMs} \\
  \hline
  \rowcolor{Gray}
  Fine-tuned \texttt{t5-large} & 52.7 $\pm$ 0.3 & 33.2 $\pm$ 0.4 & 91.9 \\
  Fine-tuned \texttt{t5-base} & 45.9 $\pm$ 0.3 & 27.6 $\pm$ 0.4 & 74.9 \\
  \rowcolor{Gray}
  Fine-tuned \texttt{t5-small} & 22.4 $\pm$ 0.3 & 11.8 $\pm$ 0.3 & 28.5 \\
  \hline
  \multicolumn{4}{c}{Baselines} \\
  \hline
  \rowcolor{Gray}
  \Gape[0pt][2pt]{\makecell[l]{Linear SVM\\(TF-IDF)}} & 48.0 $\pm$ 0.3 & 23.3 $\pm$ 0.4 & 90.2 \\
  \Gape[0pt][2pt]{\makecell[l]{Logistic Regression\\(Word counts)}} & 45.0 $\pm$ 0.3 & 23.5 $\pm$ 0.4 & 86.0 \\
  \rowcolor{Gray}
  \Gape[0pt][2pt]{\makecell[l]{Linear SVM\\(Word counts)}} & 43.6 $\pm$ 0.3 & 21.7 $\pm$ 0.4 & 88.1 \\
  \Gape[0pt][2pt]{\makecell[l]{Logistic Regression\\(TF-IDF)}} & 42.6 $\pm$ 0.3 & 19.3 $\pm$ 0.4 & 66.4 \\
  \rowcolor{Gray}
  \Gape[0pt][2pt]{\makecell[l]{Cosine Delta\\(Short Texts)}} & 24.1 & 18.2 & 89.8 \\
  Most Prominent Author & 14.2 & 4.3 & 13.2 \\
  \rowcolor{Gray}
  Random & 7.5 & 4.3 & 14.5 \\
\hline
\end{tabular}
\label{tab:methodCorr}
\end{table} 

We begin by establishing that accurate author attribution is possible for this dataset using only the available information.
It is known that authorship attribution is more reliable for longer samples.
To establish an upper bound on expected performance, we thus apply a cosine delta model to the full held-out text of each play rather than the short samples we use for all other experiments. 
This setting increases the length of attributed samples by a factor of 50 for plays in the training set and 200 for fully held-out plays.
Cosine delta accurately attributes 94.9\% of the long samples, performing better on plays in the training set than those fully held-out.

The fine-tuned \texttt{t5-large} model correctly attributes more short samples than any other method tested. 
It accurately labels 52.7\% of held-out samples from plays included in the training dataset and 33.2\% of samples from plays fully withheld from training.
\texttt{t5-large} performs substantially worse on the individual sample level than the cosine delta upper bound, but it only falls seven plays short of the upper bound when attributing plays to the most-predicted author. 
Although it is not surprising that results are better for partially-seen plays, the accuracy of both subsets exceeded our expectations.
Because the text excerpts we use are very short, they frequently contain no named entities, and we thus conclude that attribution was not performed solely using this information. 

\begin{figure}
    \centering
    \includegraphics[scale=0.35]{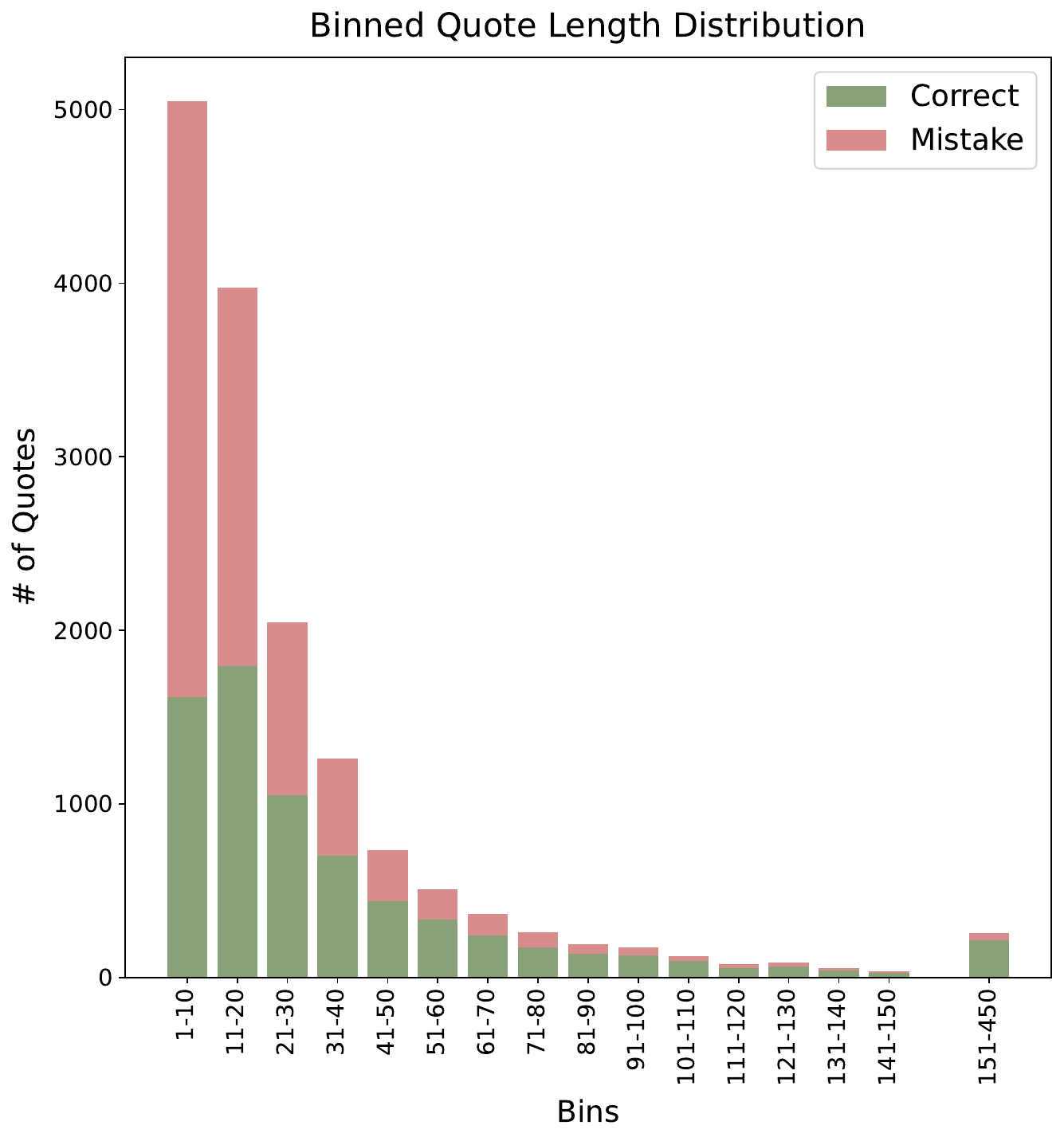}
    \caption{Length distribution of samples in the test dataset. Sample lengths are binned by 10s, and all quotes longer than 150 words are placed in one bin.}
    \label{fig:quoteLength}
\end{figure}

Longer samples were more accurately attributed. The average length of correctly attributed samples in our primary test dataset was 36.7 words whereas the average length of misattributed samples was 20.7 words.
Figure \ref{fig:quoteLength} shows the distribution of sample lengths and the accuracy for each range.
Accuracy exceeds 50\% with only 20 words (random is $\approx$5\%).
Model scale also effects accuracy.
The \texttt{t5-large} model performed better than the smaller models we compare it to, \texttt{t5-base} and \texttt{t5-small}. We observe that \texttt{t5-large} does 30.3\% better on samples from plays included in training, 21.4\% better on samples from plays withheld from training, and 63.4\% better at attributing plays by majority vote than \texttt{t5-small}.
This effect may be due to the larger model's greater capacity to fit the particulars of author-specific language in fine-tuning, a greater capacity to represent linguistic variation in pre-training, or some combination of both.

It appears that the reason for the large improvement in play attribution accuracy with model size is a significant reduction in the assignment of large numbers of samples to 2--3 specific (incorrect) authors, which we call \textit{scapegoating}. 
\texttt{t5-small} assigns 60.5\% of misattributed segments to the top two scapegoated authors (Thomas Heywood and William Shakespeare), \texttt{t5-base} assigns 32.8\% of misattributed samples to two authors (Heywood and James Shirley), and \texttt{t5-large} only attributes 25.6\% of misattributed samples to two authors (also Heywood and Shirley). Because misattributions both occur less frequently and are spread more evenly between authors in the larger models, it is more likely that the author of a play will have the majority of samples assigned to them.

Logistic regression and linear SVM prove to be strong baselines. However, \texttt{t5-large} performs 4.7\% better on samples from plays included in training and 9.9\% better on samples from withheld plays than linear SVM with TF-IDF values, the highest performing of these baselines. Since these models have access to the same data, the difference must either come from the LLM's ability to use arbitrary combinations of non-sequitive words or its access to patterns from pre-training. It is important to note that we do not know what data T5 saw during pre-training. However, because the \texttt{t5-small} model performs worse than logistic regression, it is unlikely that this is the sole source of improvement. Logistic regression and linear SVM are also prone to scapegoating: all models assign over 35\% of misattributed samples to two primary authors (Shakespeare and Shirley). Linear regression with TF-IDF values is a particularly egregious scapegoater, assigning over 50\% of samples to Shakespeare and Shirley, over double the number that \texttt{t5-large} assigns to Shirley and Heywood.

In addition to the ``merged samples'' upper bound, we apply cosine delta to the short samples. This approach performs worse than all methods but \texttt{t5-small} and the simple baselines. However, cosine delta achieves high performance at the play level. Even \texttt{t5-large} only attributes 6 more of the 253 plays in the original corpus correctly. This, again, appears to be related to scapegoating. Cosine delta assigns the samples it misattributes relatively evenly between authors, only assigning 12.4\% to the two most scapegoated authors (Richard Brome and Thomas Middleton). Thus, while cosine delta may be less accurate overall than T5, the way in which it fails is less skewed. Compared to T5, SVMs, and logistic regression, it is less likely to confidently misattribute a play.

\section{Accuracy by Author}

For the best-performing model, \texttt{t5-large}, accuracy varies considerably by author for both withheld plays and those included in training (Table \ref{tab:auCorr}). Authors with more plays in the training set are more accurately predicted for the held-out set; the Pearson correlation coefficient between these values is 0.65, with $p < 10^{-3}$.

\begin{table}
\caption{Percentage of samples correctly attributed by  \texttt{t5-large} for plays withheld from and included in the training dataset and \# of plays in the corpus by author. The authors are ordered by accuracy on samples from plays included in training.}
\small
\centering
\begin{tabular}{lrrr}
\hline
\textbf{Author} & \textbf{\protect{\% Correct (In)}} & \textbf{\protect{\% Correct (Out)}} & \textbf{\protect{\# Plays}} \\
\hline
William Shakespeare & 79.0 & 72.0 & 30 \\
\rowcolor{lightGray}
Margaret Cavendish & 74.9 & 68.5 & 12 \\
James Shirley & 61.7 & 58.5 & 31 \\
\rowcolor{lightGray}
John Lyly & 60.6 & 60.0 & 8 \\
Thomas May & 58.0 & 0.5 & 3 \\
\rowcolor{lightGray}
John Fletcher & 54.7 & 56.5 & 15 \\
Christopher Marlowe & 53.6 & 55.5 & 6 \\
\rowcolor{lightGray}
Thomas Killigrew & 52.0 & 50.0 & 4 \\
Robert Greene & 51.0 & 2.0 & 3 \\
\rowcolor{lightGray}
Ben Jonson & 49.7 & 59.5 & 14 \\
Philip Massinger & 49.0 & 46.5 & 13 \\
\rowcolor{lightGray}
Thomas Heywood & 49.0 & 36.0 & 19 \\
Richard Brome & 42.4 & 32.0 & 15 \\
\rowcolor{lightGray}
Thomas Nabbes & 40.5 & 17.5 & 5 \\
George Chapman & 38.4 & 13.0 & 11 \\
\rowcolor{lightGray}
William Davenant & 36.7 & 24.5 & 4 \\
Thomas Middleton & 36.0 & 33.0 & 13 \\
\rowcolor{lightGray}
John Marston & 34.7 & 22.5 & 7 \\
John Ford & 25.3 & 15.5 & 7 \\
\rowcolor{lightGray}
Thomas Dekker & 22.0 & 11.5 & 6 \\
Henry Glapthorne & 22.0 & 5.0 & 3 \\
\rowcolor{lightGray}
Robert Wilson & 20.0 & 19.0 & 3 \\
John Webster & 9.0 & 4.5 & 3 \\
\hline
\end{tabular}
\label{tab:auCorr}
\end{table} 

The \texttt{t5-large} model performs well on samples from many of the well-represented authors in our corpus. For 9 of the 23 authors, the model accurately attributes more than 50\% of samples from plays included in training, well above random. The four authors for whom the model performs best on samples from included plays are Shakespeare (79.0\%), Margaret Cavendish (74.9\%), Shirley (61.7\%), and John Lyly (60.6\%). The model also accurately attributes many samples from the withheld plays by these authors: 72.0\% of samples from Shakespeare\footnote{\textit{Antony and Cleopatra}} are correctly attributed, 68.5\% of samples from Cavendish\footnote{\textit{The Wooers}}, 58.5\% of samples from Shirley\footnote{\textit{The Sisters}}, and 60.0\% of samples from Lyly\footnote{\textit{Sappho and Phao}}.

The reasons that the model attributes samples from these four authors with such high accuracy differ. Shirley and Shakespeare are the authors with the most and second-most plays in the dataset, with 31 and 30 plays in the corpus respectively. But Cavendish (8) and Lyly (12) are close to the average. Authors comparable to Cavendish in representation, such as George Chapman (11), Philip Massinger (13), and Thomas Middleton (13), all have accuracies below 50\% for plays included in training. Similarly, authors comparable to Lyly such as John Ford (7) and John Marston (7) both have accuracies below 35\% on included plays. Therefore, there is likely something distinctive about these two authors that makes them easier for the model to identify. Note that Cavendish is the only female author in the corpus (we were unable to include others), so we are not able to determine if her plays are distinctive because she has an individual style or if women authors of the period wrote differently from men.

To further explore the cause of Cavendish and Lyly's distinctiveness, we compare each author's usage of the 100 most frequent words in the corpus. We first calculate z-scores comparing the frequency with which an author used each word to the mean frequency of that word's usage for all authors in the dataset. The frequencies are normalized by author so that no single author skews the distribution and we ensure that the set of 100 most frequent words contains no named entities. We then sum the absolute values of each author's z-scores to create a `uniqueness' metric. For a further exploration and validation of this metric, please see Section B of the appendix.
The summed z-scores ranged from 47.5 to 138.2. The author with the most unique usage of common words by this metric is Cavendish, with a score of 138.2. The authors with comparable play counts to Cavendish each have considerably lower scores (Chapman: 50.57, Massinger: 69.9, Middleton: 67.7). The second most distinctive author is Thomas Killigrew, with a summed z-score of 108.7. Indeed, Killigrew has a very high accuracy on samples from included plays (52.0\%) considering only four of his works are in the corpus. Lyly also has a relatively high summed z-score of 99.4, which is the fourth largest in the dataset. Again, this is higher than the scores of comparably represented authors (Ford: 68.3, Marston: 70.8), but not by as much. Notably, both Shirley and Shakespeare have low uniqueness scores by this metric. Shakespeare's is the lowest (47.5) and Shirley's is the 16th lowest (67.8). In addition, both authors have large vocabularies; Shakespeare has the largest vocabulary and Shirley the third-largest of all authors in the dataset. Both of these trends are likely related to their prominence within the training dataset, but they may still be meaningful. It is possible that Shakespeare and Shirley's uniqueness comes from using words that the other authors do not, instead of using common words uniquely. Overall, it seems that an author's usage of common words does affect how well the model can identify their writing. But it does not explain all of the variation seen in the dataset.

\subsection{Quote Misattribution}

\begin{figure}[ht]
    \centering
    \includegraphics[scale=0.4]{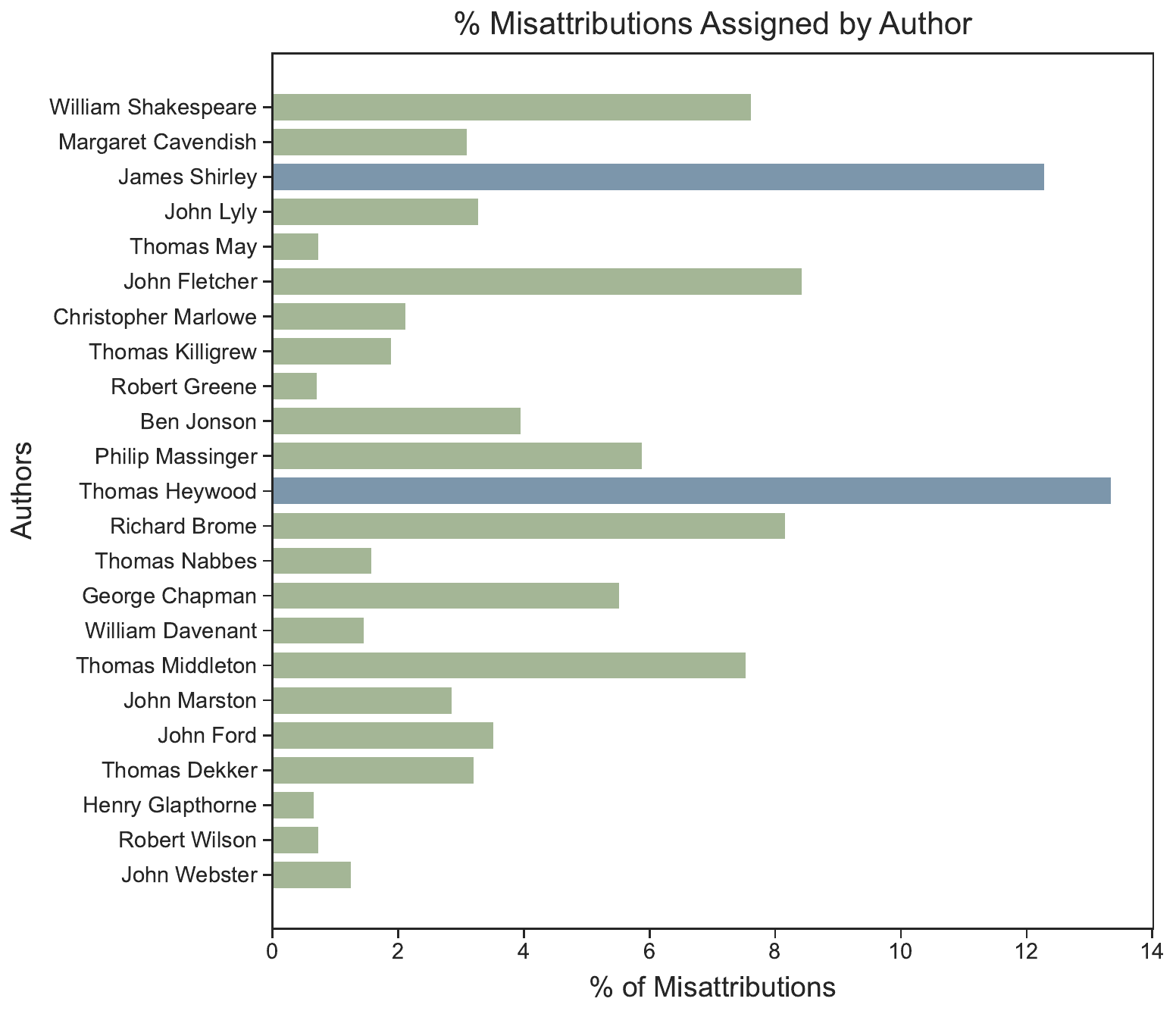}
    \caption{Percentage of misattributed samples assigned to each author.}
    \label{fig:wrongByAu}
\end{figure}

\begin{figure}[ht]
    \centering
    \includegraphics[scale=0.25]{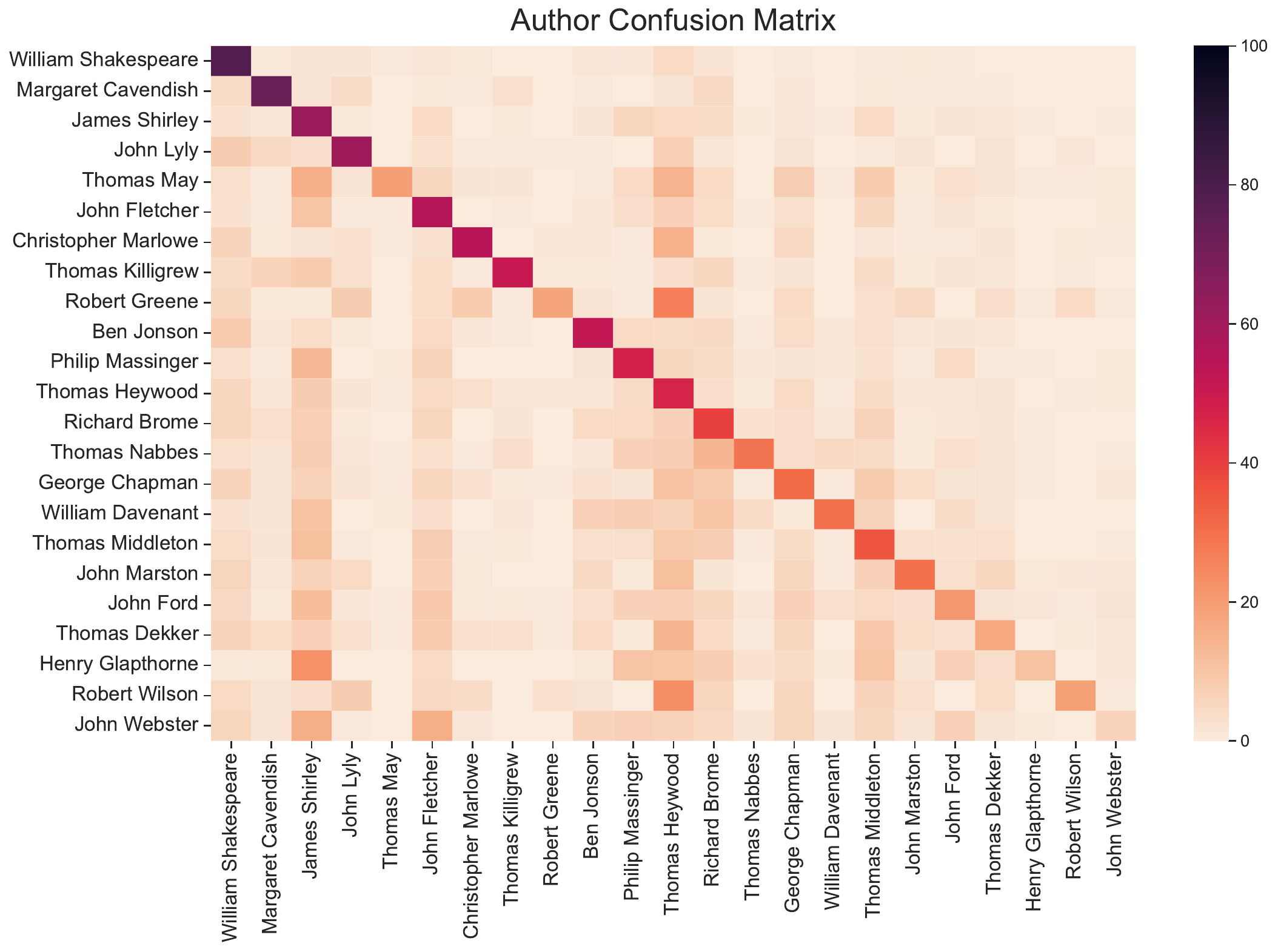}
    \caption{Confusion matrix demonstrating how frequently samples from row authors were misattributed to column authors. Each matrix row sums to 100\%. Prolific authors Heywood, Shakespeare, and Shirley are most commonly guessed.}
    \label{fig:confMatrix}
\end{figure}

\begin{figure}[ht]
    \centering
    \includegraphics[scale=0.4]{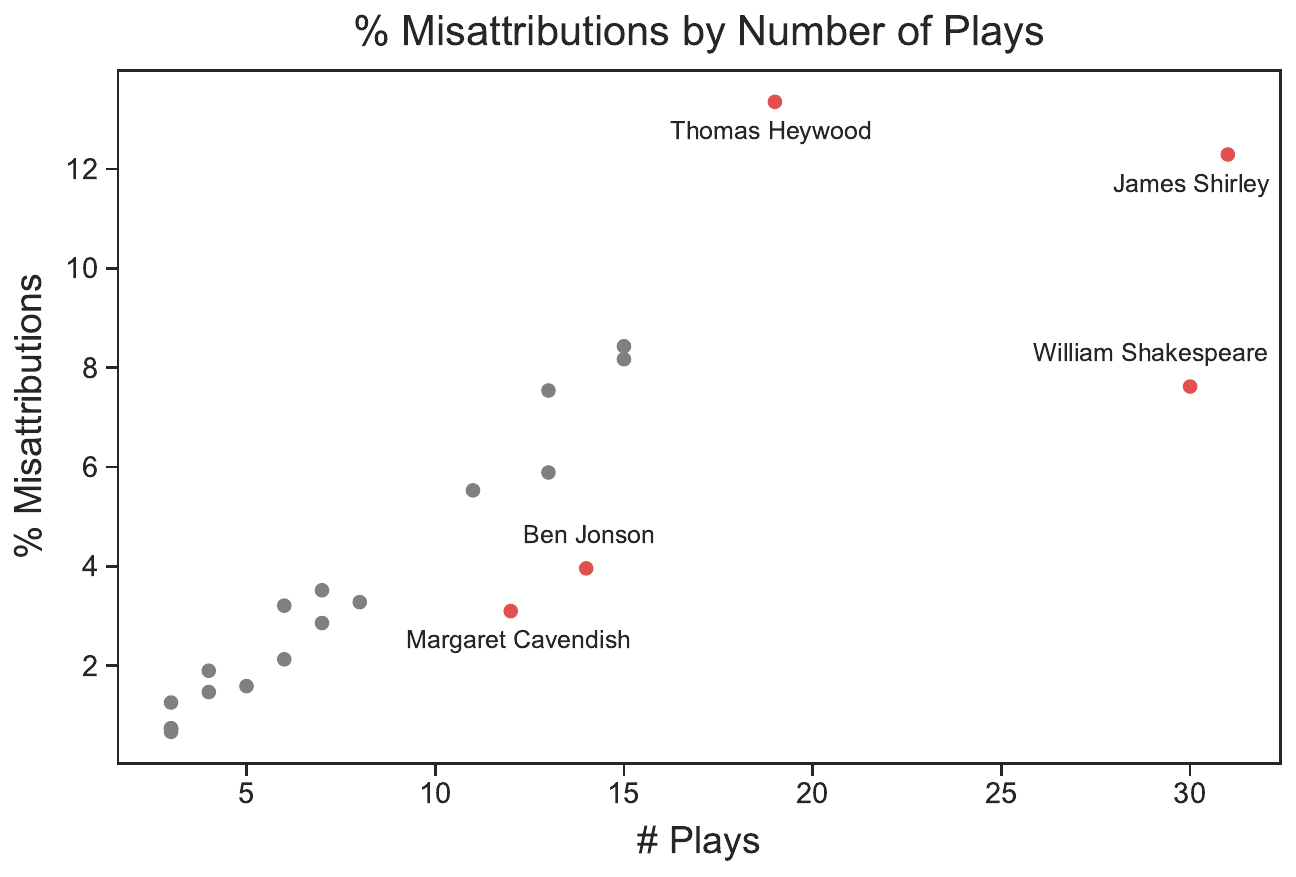}
    \caption{The relationship between the number of plays by an author and the \% of misattributed samples assigned to them. Visual outliers are highlighted and labeled.}
    \label{fig:wrongByNumPlays}
\end{figure}

There is also considerable variation in how the fine-tuned \texttt{t5-large} model misattributes quotes (Figure \ref{fig:wrongByAu}). Instead of assigning the misattributed quotes to authors randomly, it scapegoats two primary authors, Heywood and Shirley, and assigns them a disproportionate number. A confusion matrix depicting who quotes are misattributed to by original author demonstrates that the scapegoating phenomenon is not caused by confusion between specific pairs of authors (Figure \ref{fig:confMatrix}). Instead, the misattributions to Heywood and Shirley are spread throughout the dataset. Again, it appears that contribution to the corpus is one factor that affects who samples are misattributed to. The Pearson's R correlation between the number of plays by an author in the dataset and the percentage of misattributed samples assigned to them is 0.86 with $p < 10^{-6}$. The outliers from this relationship appear to be Heywood, Shakespeare, Cavendish, and Ben Jonson (Figure \ref{fig:wrongByNumPlays}).

Examining authors' scores for the summed z-score metric again provides an indication of why some are scapegoated. Cavendish's high uniqueness score likely means it is more difficult for the model to mistake a given quote for hers. In contrast, Heywood, who has the most samples misattributed to him, has the second-lowest uniqueness score in the corpus, 49.4. He also has the second-largest vocabulary. The combination of these factors may help explain why he is so frequently scapegoated. Given a random quote from the test dataset, Heywood is more likely than most authors to have all of the words in the sample in his vocabulary. Even if he doesn't, the model could have learned that he is more likely to use a broad range of words than other authors. In addition, common word usage in the average corpus sample is likely to resemble Heywood's. Shirley, who has the second-most misattributed quotes assigned to him, has the third-largest vocabulary and the 16th lowest uniqueness score. Thus, it appears that vocabulary size and common word usage are factors that affect to whom the model's misattributes quotes.

However, there are two major outliers which indicate that these three factors---number of plays, common word usage, and vocabulary size---cannot be the only ones affecting scapegoating. These are Jonson and Shakespeare. Shakespeare has both the largest vocabulary and lowest uniqueness score of any author in the corpus, and yet samples are less likely to be misattributed to him than would be expected given his contribution to the dataset. Similarly, Jonson has the fourth-largest vocabulary and the 19th lowest uniqueness score, yet he also stands out as an outlier to whom fewer samples are misattributed than expected. We hypothesize that these outliers are caused by the model's pre-training. Of the authors included in our corpus, Shakespeare and Jonson are among the best-known today. The model is likely to have seen the writing of these authors during pre-training, and may therefore be more likely to correctly label data from these authors than would be expected given only the fine-tuning process.

\section{Accuracy by Play}

Interesting patterns and outliers emerge when we examine the model's play-by-play accuracy at attributing samples. There are several authors, like Cavendish, for whom the proportion of correctly attributed samples is largely consistent across all plays and some, like Brome, for whom there is considerable variation in the play-level accuracy but for whom there are no noticeable outliers. When there is an outlier among an author's plays, there is usually (though not always) an identifiable reason for why that play stylistically differs from the rest of the author's work. 

\begin{figure}[ht]
    \centering
    \includegraphics[scale=0.35]{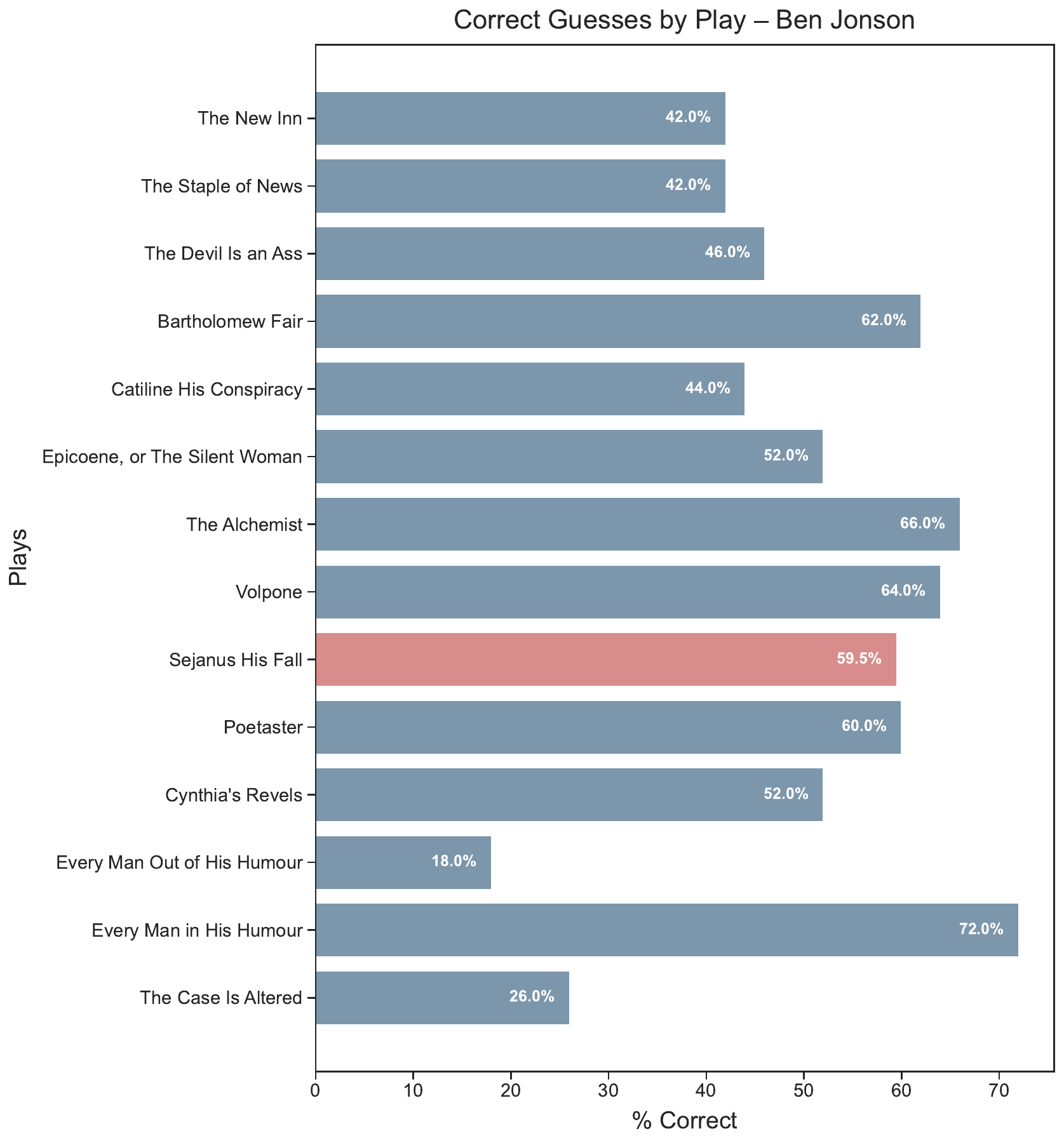}
    \caption{Percentage of samples correctly attributed for each play by Ben Jonson. The pink bar represents the play withheld from training.}
    \label{fig:jonson}
\end{figure}

A representative example of this can be seen in Ben Jonson's plays. Samples from all but two of Jonson's plays are correctly attributed more than 40\% of the time, including those from the withheld play (Figure \ref{fig:jonson}). However, only 18\% of samples from \textit{Every Man Out of His Humour} and 26\% of samples from \textit{The Case is Altered} are attributed to Jonson, causing \textit{Every Man Out of His Humour} to be misattributed by the model. Both of these plays differ from Jonson's typical work. Although \textit{Every Man Out of His Humour} was advertised as a sequel to the well-received \textit{Every Man in His Humour}, it is very different from the original play \citep{ondb}. It was the longest play written for a public theater performance during the Elizabethan era and was very poorly received. After its failure, Jonson began writing for private theaters instead \citep{augustyn2014}. Thus, it is likely that the play marks a stylistic experiment within Jonson's work. Interestingly, this play is still correctly attributed by sample-level cosine delta with 16\% of samples. \textit{The Case is Altered} is unique because it is the earliest surviving of Jonson's plays. Jonson excluded it from his collected works when they were first published and, even when it was eventually published in 1609, his name was only included in some copies \citep{oliphant1911}. \textit{The Case is Altered} therefore likely represents an early work which the author was not proud of, and from whose style he matured.

\begin{figure}[ht]
    \centering
    \includegraphics[scale=0.35]{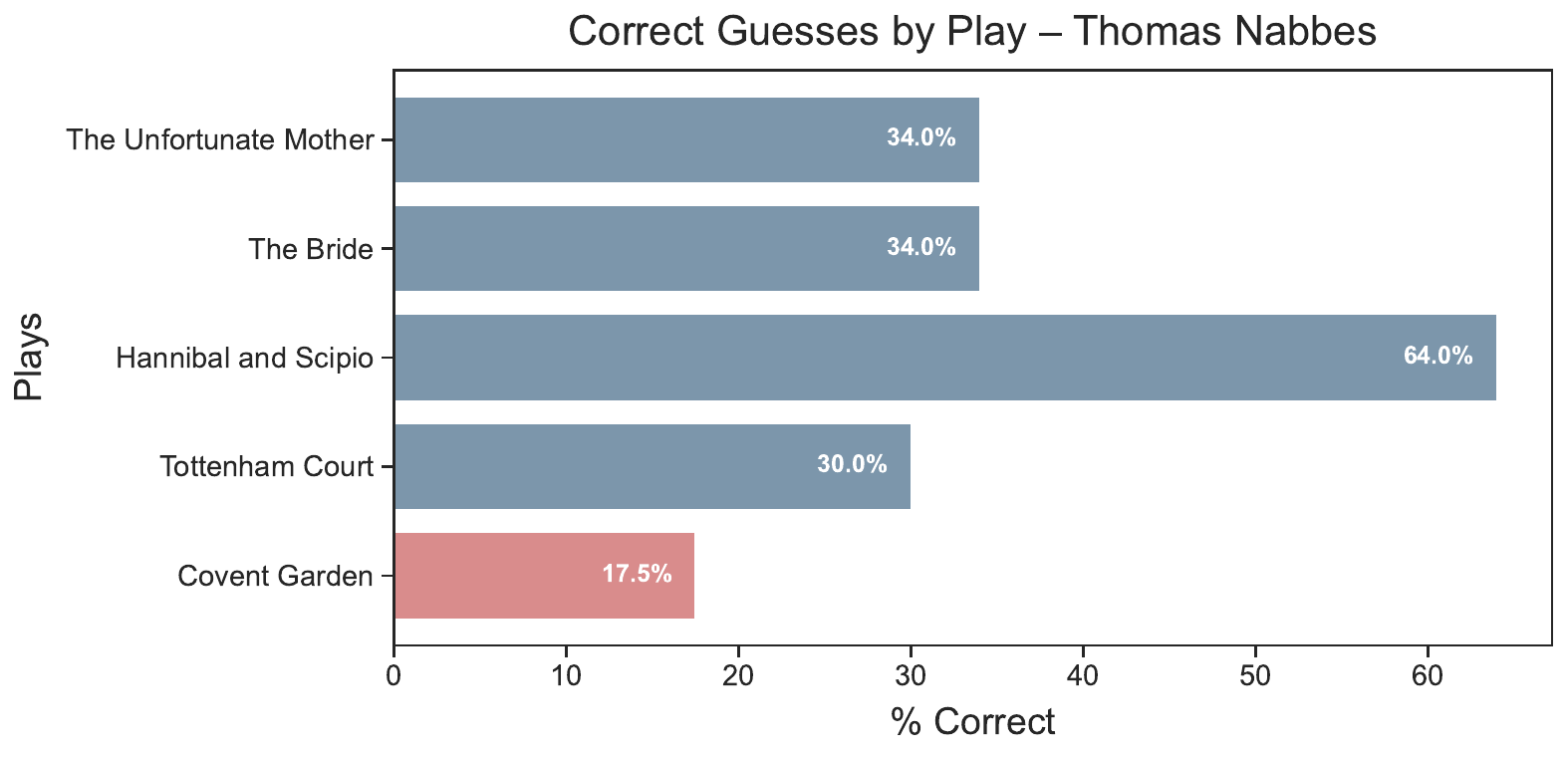}
    \caption{Percentage of samples correctly attributed for each play by Thomas Nabbes. The pink bar represents the play withheld from training.}
    \label{fig:nabbes}
\end{figure}

Another outlier is \textit{Covent Garden} by Thomas Nabbes. Although there is some play-level variation in Nabbes' attribution accuracy, \textit{Covent Garden} is the only play for whom the model correctly attributes less than 20\% of samples and the only one it misattributes, assigning Brome 23\% of samples (Figure \ref{fig:nabbes}). While this variation may be in part because \textit{Covent Garden} was withheld from training, the underlying reason for the model's confusion is likely that Nabbes' \textit{Covent Garden} was written as a direct response to Richard Brome's play \textit{The Weeding of Covent Garden}, which is also included in the dataset. There are likely named entities that cross-over between these two works and there may even be stylistic similarities. Sample-level cosine delta correctly attributes \textit{Covent Garden} to Nabbes, but with only 13\% of samples. It assigns 10\% of samples to Brome.

We also see that the model performs poorly on the withheld plays of almost all authors with only three works in the corpus. For four of these five authors, less than 5\% of samples from withheld plays are correctly attributed. The only deviation from this pattern is Robert Wilson; 19\% of samples from the withheld Wilson play are correctly attributed. However, the withheld Wilson play is a prequel to one included in the training set. Thus, the model has more knowledge of this play than it would otherwise. It appears that including two plays by an author, or 470 samples, in the training data is not sufficient for the model to learn to extrapolate an author's style to an unseen text. It thus suggests a boundary for how much data may be needed for LLMs to be used for authorship attribution. 

\subsection{Disputed and Co-Authorship}

We also asked the \texttt{t5-large} model predict the author of samples from 23 plays which are of disputed authorship or which are believed to be co-authored, although they were labeled as written by a single author in the corpora we drew from. We determined which plays were co-authored or of disputed authorship using the Oxford National Dictionary of Biography, which provides a detailed biography for each author in this corpus. 

\begin{wrapfigure}{l}{0.65\textwidth}
    \centering
    \includegraphics[scale=0.35]{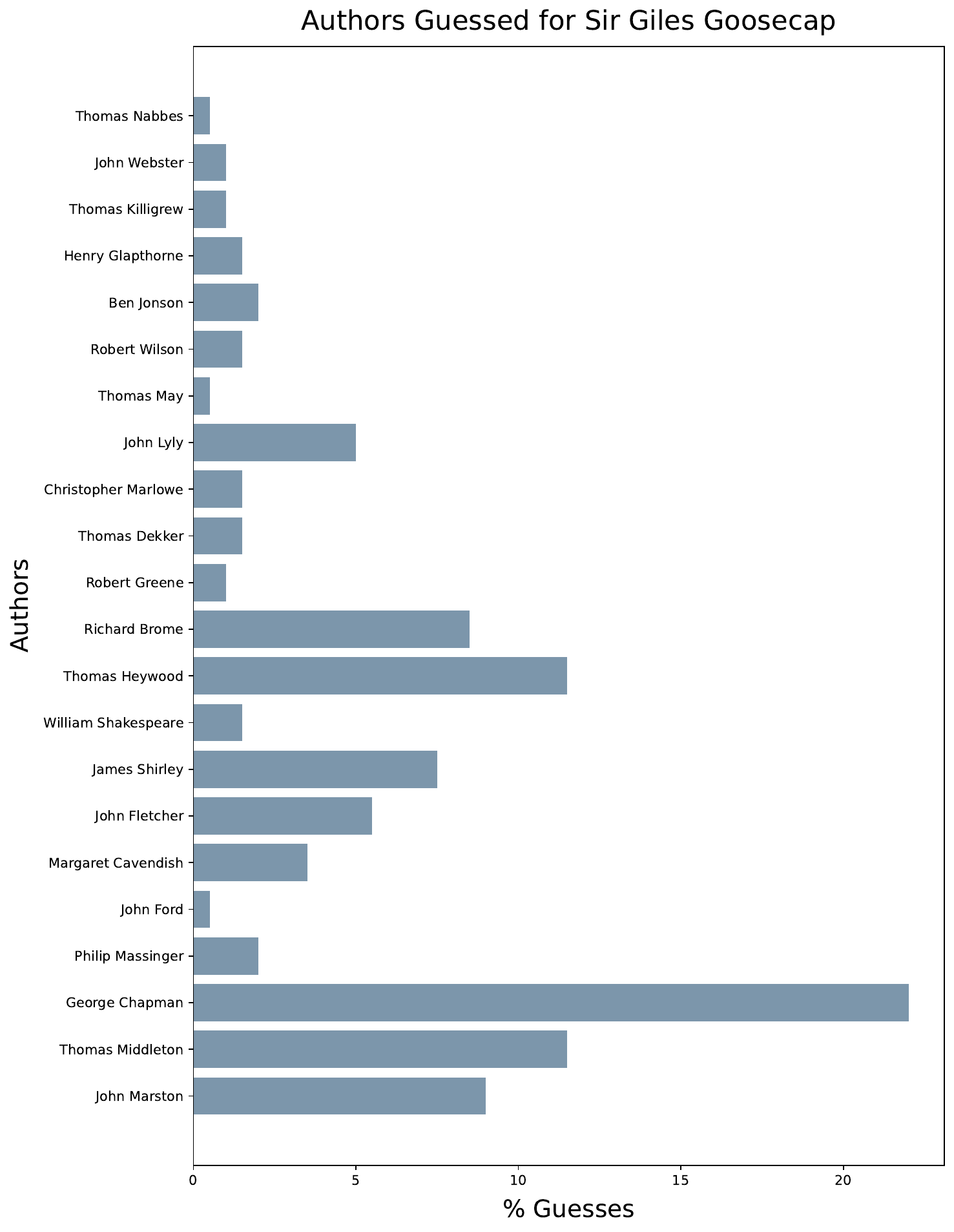}
    \caption{Percentage of samples attributed to authors for \textit{Sir Giles Goosecap}.}
    \label{fig:sirGiles}
\end{wrapfigure}

Overall, we found that the model greatly struggles to make clear attributions for plays that were co-authored or of disputed authorship unless Shakespeare was a contributor. The only exception is \textit{Sir Giles Goosecap}, which is hypothesized to have been written by George Chapman. The model attributes 22\% of samples from \textit{Sir Giles Goosecap} to Chapman (Figure \ref{fig:sirGiles}). This is comparable to two other plays by Chapman in the original dataset: \textit{All Fools}, which was withheld from training and from which 13\% of samples are correctly attributed, and \textit{The Widow's Tears}, from which 24\% of samples are correctly attributed. Thus, the model results support the overall attribution of this play to Chapman. Sample-level cosine delta does not support this attribution, assigning only 5.5\% of samples to Chapman. For no other non-Shakespearian play in this subcorpus is there enough evidence to argue for an attribution or co-attribution to authors in the dataset. It is particularly difficult to make assumptions about plays that are suspected to have been written by authors for whom the model's performance on the initial corpus is low. Even if these authors are only attributed a small proportion of samples from a play, these results are often comparable to those for their plays in the original dataset, meaning no conclusion can be reached. 

Co-authorship also confused the model, particularly plays that were co-written with authors outside of the original corpus. The model often attributed a large proportion of quotes from these plays to Heywood. However, since there is no evidence that Heywood helped to author these plays, it is likely that this is an artifact of scapegoating. This trend also means that it is difficult to attribute plays to Heywood. 22\% of samples from \textit{The Fair Maid of the Exchange}, which Heywood is suspected to have co-authored, are attributed to him. However, a comparable proportion of samples are attributed to Heywood for multiple other plays in this dataset, meaning that we cannot use this as evidence for his authorship. This is a clear example of a case in which the model's misattribution patterns detrimentally affect its usability. The results are confusing even for plays from whom all of the suspected contributors are in the dataset, such as \textit{The Laws of Candy}.

Thus, the results for non-Shakespearian plays provide little evidence for or against certain writers' authorship. While sample-level cosine delta appears to have no clear advantage over \texttt{t5-large} in attributing samples from these plays, the two methods attribute samples in very different ways. In some cases, \texttt{t5-large} more strongly attributes a play to its suspected author, and in others sample-level cosine delta does. Often the models attributed samples to different subsets of authors.

A very interesting pattern emerges when we look at the plays co-authored by Shakespeare in this test corpus. Over 50\% of samples from each of the eight plays that Shakespeare contributed to are attributed to him, with little to no samples attributed to those who he supposedly co-authored the plays with, even if they are in the dataset. The most significant indication we see of another author's contribution to one of these plays is for \textit{The Two Noble Kinsmen}. Here, only 54\% of samples are attributed to Shakespeare and 8.5\% are attributed to Fletcher, with whom he wrote the play. However, this is still not a strong signal of Fletcher's involvement. This pattern again suggests that the \texttt{t5-large} model recognizes Shakespeare from pre-training. If the model had seen these plays attributed solely to Shakespeare during pre-training, as is likely, it may help explain why it assigns them so confidently to Shakespeare despite the influence of other authors. In contrast, sample-level cosine delta never assigns more than 25\% of samples from any of these plays to Shakespeare, and the presence of his theorized co-authors is much more prominent in the results.

\section{Stylistic Development Over Time}
\label{sec:style-over-time}

\begin{table}[b]
\caption{The number of plays and authors by century in the comparison dataset.}
\small
\centering
\begin{tabular}{lrrrrr}
\hline
\textbf{Century} & \textbf{\protect{\# Plays}} & \textbf{\protect{\# Authors}} \\
\hline
1500s & 14 & 13 \\
1600s & 61 & 53 \\
1700s & 18 & 16 \\
1800s & 18 & 14 \\
1900s & 29 & 13 \\
\hline
\end{tabular}
\label{tab:styleShift}
\end{table} 

\begin{figure}[ht]
    \centering
    \includegraphics[scale=0.4]{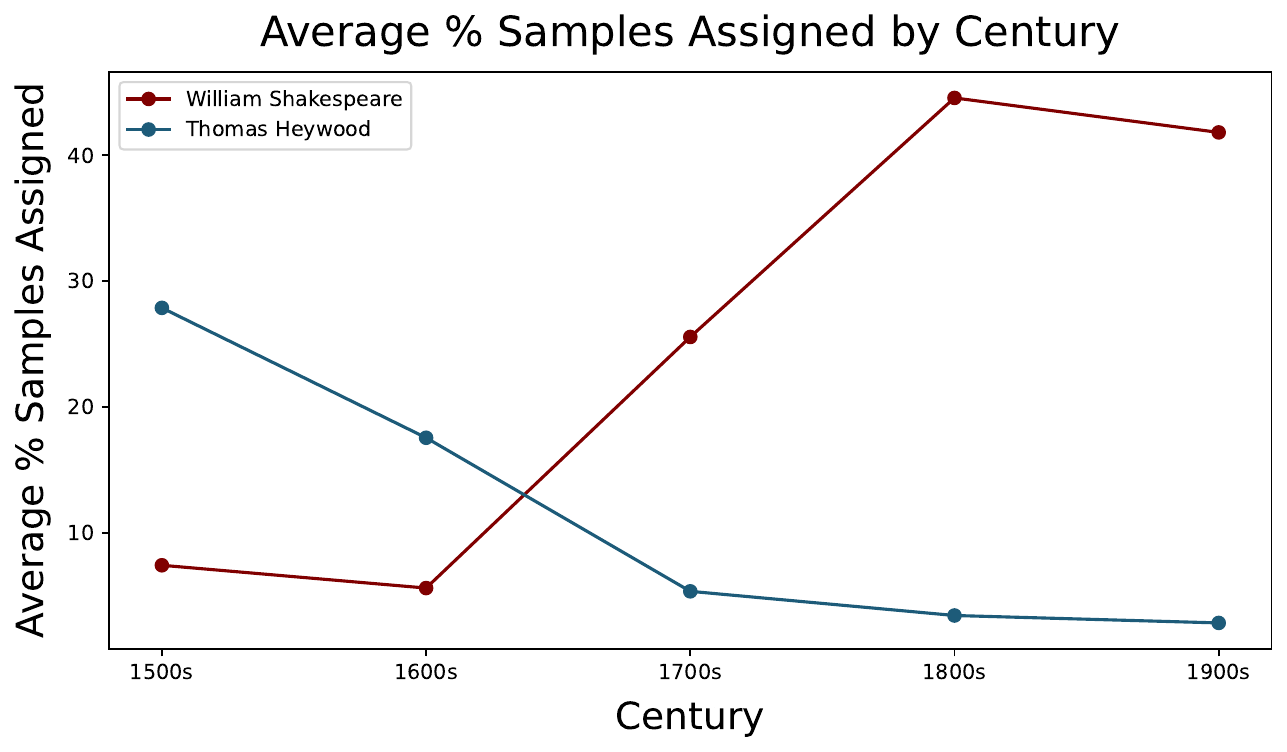}
    \caption{The average \% of samples assigned to Thomas Heywood and William Shakespeare for samples from each century in the comparative corpus.}
    \label{fig:styleByTime}
\end{figure}

In addition to the narrower task of author attribution, a measure of stylometric similarity can also be used to quantify authors' influence. To study shifts in dramatic style over time, we created a comparative corpus of plays written between the 14th and 18th centuries. In this corpus, we included 74 plays gathered from the EMED and SHC corpora not written by authors in our training dataset. To these, we added 67 additional plays from Project Gutenberg (see Table \ref{tab:styleShift}). We performed the same utterance separation and splitting with these plays as with the original corpus and formatted the input and output pairs identically. Further details can be found in the appendix. The \texttt{t5-large} model fine-tuned on the original dataset was asked to predict authors for 200 samples from each of these plays. The percentages we report in Figure \ref{fig:styleByTime} are averaged by the original text author instead of by play; for example, we calculate the percentage of samples attributed to Heywood from each author writing in the 1500s and then average those percentages to reach the depicted value. This was to prevent any single writer whose style may somehow mimic that of an author in our original dataset from skewing the results. 

In the 1500s and 1600s, the greatest proportion of samples are assigned to Thomas Heywood. This aligns with the scapegoating trends we saw in the original corpus. However, starting in the 1700s the greatest proportion of samples are assigned to Shakespeare, and this value increases in the 1800s and 1900s (Figure \ref{fig:styleByTime}), for which nearly half of the samples from each author were attributed to Shakespeare. In addition, if we attribute plays to an author by majority vote, no plays in the 1500s are assigned to Shakespeare, but 97\% of plays are attributed to him by the 1900s. This result does not imply that 20th century plays are similar to Shakespeare, only that of the Early Modern authors known to the model, Shakespeare is both distinct and increasingly more similar to more recent plays than any other Early Modern author.

\section{Conclusion}

Generative large language models provide a promising tool for stylometry.
While simpler methods such as cosine delta remain more accurate for larger text segments, we find that LLMs, particularly at larger scales, are remarkably effective at predicting the author of a difficult corpus of short 5--450 word text segments, which are more aligned with LLMs' shorter input windows.

In addition to quantitative power, LLM-based stylometric analysis provides evidence for a range of interpretive arguments both when it succeeds (such as with Margaret Cavendish) as well as when it fails (both in scapegoating and in the stylistic differences in the work of Ben Jonson). Because T5 demonstrates an ability to recognize style, it may prove useful in other situations where recognizing implicit signals is key such as tracking genre differences and stylistic movements. 
There are also substantial practical advantages to using fine-tuned LLMs: despite their complexity and computational intensity, generative LLMs provide a remarkably simple text-in/text-out user interaction that requires no specialized software.

However, there are several disadvantages to using pre-trained LLMs for authorship attribution. They are more computationally intensive than more traditional methods of authorship attribution and the content and effect of pre-training corpora are difficult to assess. In addition, the ways in which the model confidently misattributes texts means that it is more likely to produce misleading results than traditional attribution methods. Given the differences that emerged between the performance of cosine delta and the fine-tuned LLM, using the two methods in conjunction may provide more accurate results than using either method separately. Due to the weaknesses we have observed, however, we recommend against using LLMs for authorship attribution in forensic or legal settings.

\begin{acknowledgments}
  We would like to thank Federica Bologna, Katherine Lee, Noam Ringach, Rosamond Thalken, Andrea Wang, Matthew Wilkens, and Gregory Yauney for their thoughtful feedback. This work was supported by the NEH project AI for Humanists and Cornell University's Hopcroft Fellowship.
\end{acknowledgments}

\bibliography{main}
\pagebreak
\appendix

\section{T5 Fine-Tuning Hyperparameters}

\renewcommand{\arraystretch}{1.5}
\begin{longtable}{|p{0.5\linewidth} | p{0.2\linewidth}|}
\hline
\textbf{Parameter} & \textbf{Value} \\
\hline
Evaluation Strategy & Epoch \\
\hline
Learning Rate & 2x$10^{-5}$ \\
\hline
Weight Decay & 0.01 \\
\hline
Save Total Limit & 3 \\
\hline
\end{longtable}

\section{Examination of Z-Score Uniqueness}

\begin{figure}[b]
    \centering
    \includegraphics[scale=0.4]{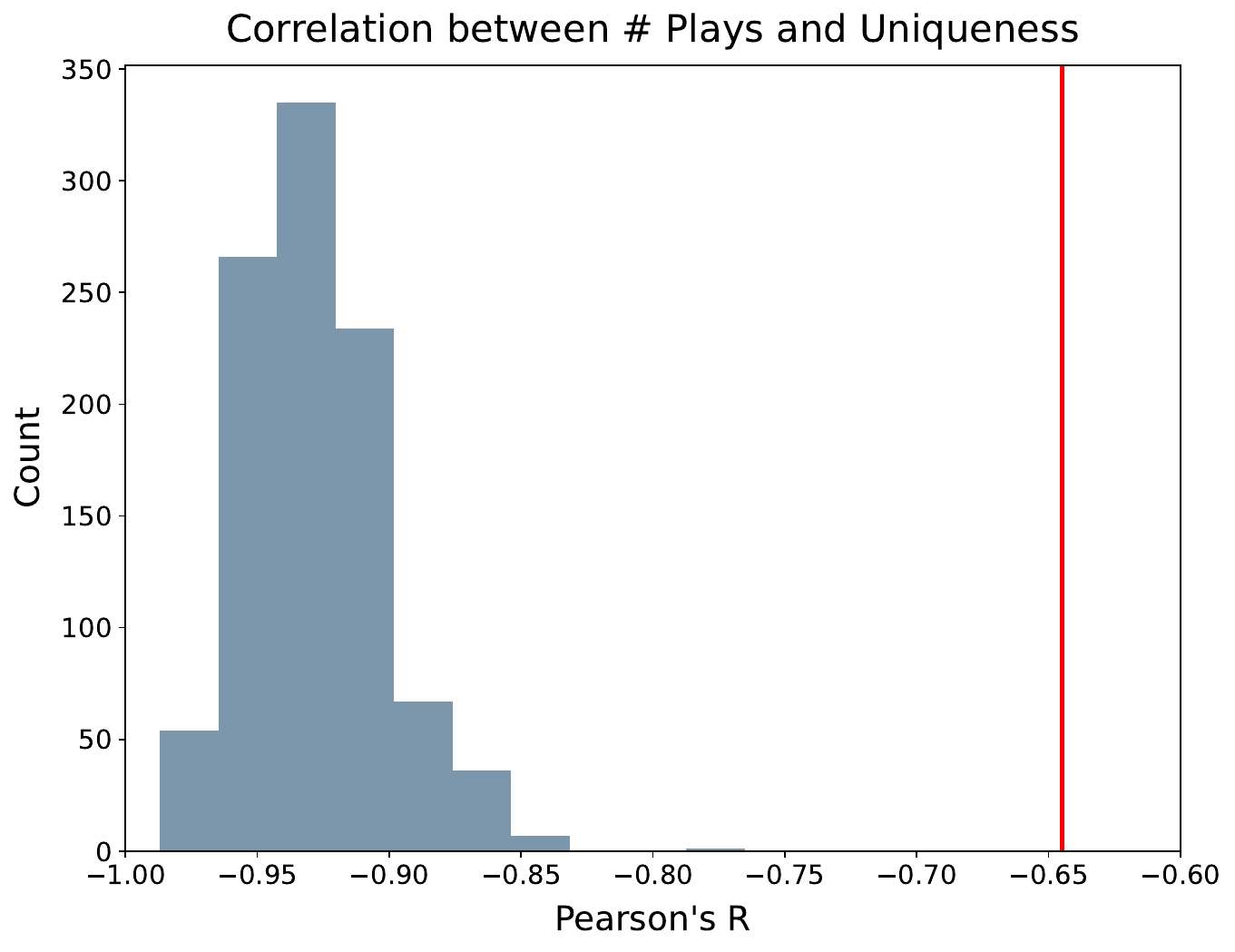}
    \caption{Binned correlations between uniqueness scores and number of plays from each synthesized trial. The red line represents the correlation from our true dataset.}
    \label{fig:uniqueCorr}
\end{figure}

To explore the validity of our uniqueness metric, we ran 1,000 synthesized trials to examine what the expected correlation between dataset contribution and the uniqueness metric would be given randomly assigned plays. Concretely, in each trial we randomly assigned plays to synthetic authors in the same proportions they are assigned to authors in our true dataset. We then calculated the Spearman's rho correlation between number of plays and uniqueness values for each trial. We plot the binned synthetic correlations and the true correlation from our dataset in Figure \ref{fig:uniqueCorr}. The true correlation from our dataset, depicted with the vertical red line, is 0.2 away from any value reached in our synthesized trial. Thus, it seems that there are some notable deviations from the expected trend in our dataset.

\begin{figure}
    \centering
    \includegraphics[scale=0.5]{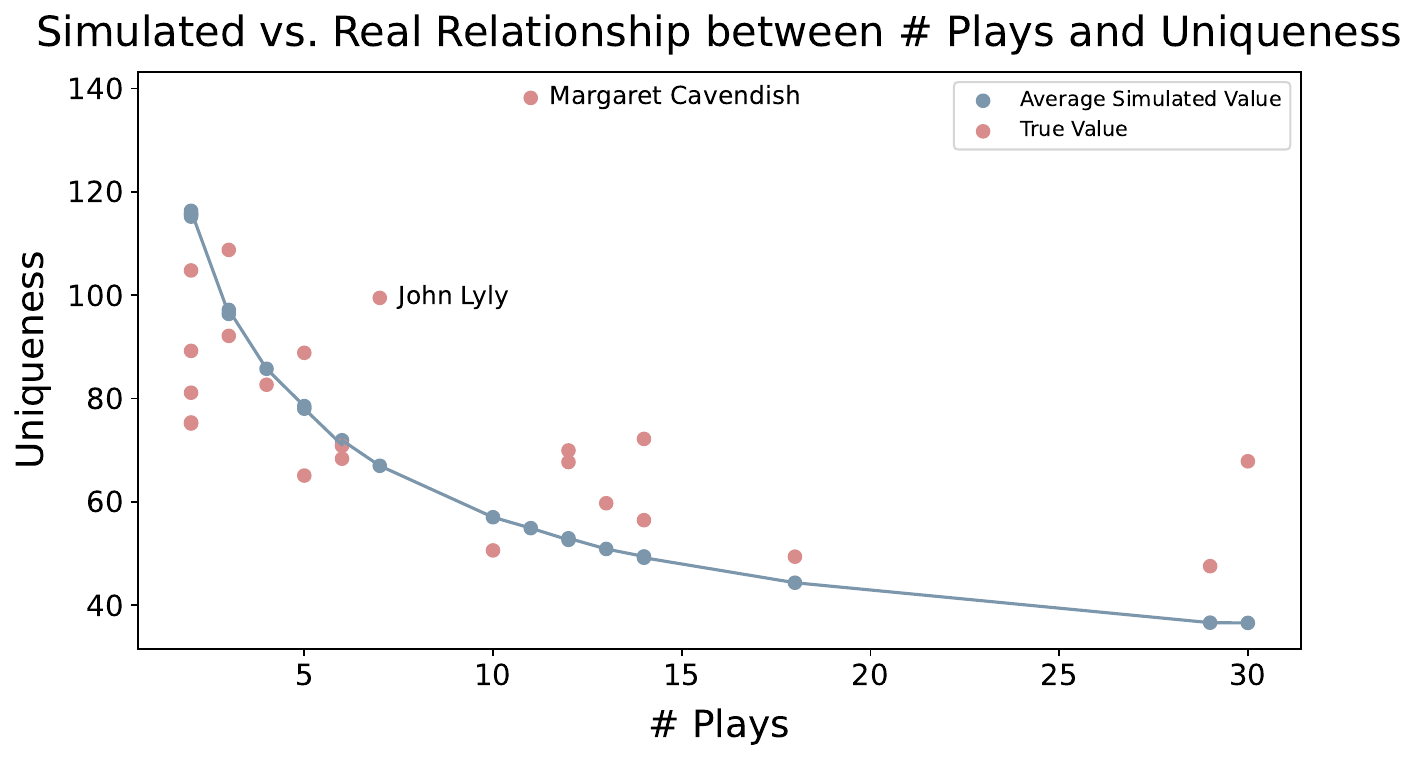}
    \caption{The relationship between the averaged uniqueness values for each synthetic author (blue) and the number of plays and the true uniqueness values from the corpus (red) and the number of plays.}
    \label{fig:uniquePlot}
\end{figure}

To further explore this relationship, we averaged the uniqueness values for each synthetic author over all trials and plotted these values as well as the true values in Figure \ref{fig:uniquePlot}. It is clear that the true uniqueness values frequently deviate from the expected relationship between uniqueness and number of plays. In particular, Margaret Cavendish and John Lyly have much higher uniqueness values than expected given the number of plays they contribute to the training dataset. Because of this, we believe that this metric represents a valuable measure of uniqueness and does not simply reemphasize the impact of contribution to the training corpus.

\section{Original Corpus Contents}

All plays in our original training and test corpora by author. The withheld plays are bolded and italicized.

\renewcommand{\arraystretch}{1.5}
\begin{longtable}{|p{0.2\linewidth} | p{0.75\linewidth}|}
\hline
\textbf{Author} & \textbf{Plays} \\
\hline
Richard Brome & The Northern Lass, The City Wit or The Woman Wears the Breeches, The Queen's Exchange (The Royal Exchange), The Weeding of Covent Garden or The Middlesex Justice of Peace, The Novella, The Queen and Concubine, The New Academy or The New Exchange, The Sparagus Garden (Tom Hoydon o' Tanton Deane), \textbf{\textit{The English Moor or The Mock Marriage}}, The Antipodes, The Damoiselle or The New Ordinary, A Mad Couple Well Matched, The Lovesick Court or The Ambitious Politic, The Court Beggar, A Jovial Crew or The Merry Beggars\rule{0pt}{1em}\\
\hline
Margaret Cavendish & The Lady --- Part 1, The Lady --- Part 2, The Unnatural Tragedy, Wit's Cabal --- Part 1, Wit's Cabal --- Part 2, Love's Adventures --- Part 1, Love's Adventures --- Part 2, Several Wits, The Matrimonial Trouble --- Part 1, The Matrimonial Trouble --- Part 2, The Religious, \textbf{\textit{The Wooers}} \\
\hline
George Chapman & The Blind Beggar of Alexandria, A Humorous Day's Mirth, \textbf{\textit{All Fools}}, The Gentleman Usher, May Day, The Widow's Tears, Bussy D'Ambois, Monsieur D'Olive, Caesar and Pompey (The Wars of Caesar and Pompey), The Tragedy of Charles Duke of Byron, The Revenge of Bussy D'Ambois \\
\hline
William Davenant & The Cruel Brother, Albovine King of the Lombards, The Just Italian, \textbf{\textit{The Wits}} \\
\hline
Thomas Dekker & Old Fortunatus, Satiromastix or The Untrussing of the Humorous Poet, \textbf{\textit{The Honest Whore --- Part 2}}, Match Me in London, If It Be Not Good the Devil Is in It \\
\hline
John Fletcher & The Faithful Shepherdess, The Woman's Prize or The Tamer Tamed, Bonduca, Valentinian, The Mad Lover, The Chances, The Loyal Subject, The Humorous Lieutenant (Generous Enemies, Demetrius and Enanthe), Women Pleased, \textbf{\textit{The Island Princess}}, The Wild Goose Chase, The Pilgrim, Rule a Wife and Have a Wife, A Wife for a Month \\
\hline
John Ford & The Lover's Melancholy, The Broken Heart, 'Tis Pity She's a Whore, \textbf{\textit{Love's Sacrifice}}, Perkin Warbeck, The Fancies Chaste and Noble \\
\hline
Henry Glapthorne & The Hollander, Ladies' Privilege, \textbf{\textit{Wit in a Constable}} \\
\hline
Robert Greene & Friar Bacon and Friar Bongay, \textbf{\textit{The Scottish History of James the Fourth}}, Orlando Furioso \\
\hline
Thomas Heywood & The Four Prentices of London, Edward IV --- Part 1, Edward I --- Part 2, \textbf{\textit{The Royal King and the Loyal Subject}}, How a Man May Choose a Good Wife from a Bad, A Woman Killed with Kindness, If You Know Me Not You Know Nobody or The Troubles of Queen Elizabeth --- Part 1, If You Know Me Not You Know Nobody or The Troubles of Queen Elizabeth --- Part 2, The Fair Maid of the West or A Girl Worth Gold --- Part 1, The Wise Woman of Hogsdon, The Rape of Lucrece, The Golden Age or The Lives of Jupiter and Saturn, The Brazen Age, The Iron Age --- Part 1, The Iron Age --- Part 2, The English Traveller, Love's Mistress, A Challenge for Beauty \\
\hline
Ben Jonson & The Case is Altered, Every Man in His Humour, Every Man Out of His Humour, Cynthia's Revels, Poetaster, \textbf{\textit{Sejanus His Fall}}, Volpone, The Alchemist, Epicoene or The Silent Women, Catiline His Conspiracy, Bartholomew Fair, The Devil is an Ass, The Staple of News, The New Inn \\
\hline
Thomas Killigrew & The Prisoners, The Princess, The Parson's Wedding, \textbf{\textit{Claricilla}} \\
\hline
John Lyly & \textbf{\textit{Sappho and Phao}}, Campaspe (Alexander, Campaspe, and Diogenes), Gallathea, Endymion, Midas, Love's Metamorphosis, Mother Bombie, The Woman in the Moon \\
\hline
\makecell[l]{Christopher\\Marlowe} & \textbf{\textit{Tamburlaine the Great --- Part 1}}, Tamburlaine the Great --- Part 2, The Jew of Malta, Doctor Faustus, Edward the Second, THe Massacre at Paris \\
\hline
John Marston & Antonio and Mellida, Antonio's Revenge, Jack Drum's Entertainment, What You Will, \textbf{\textit{The Malcontent}}, Parasitaster or The Fawn, The Dutch Courtesan \\
\hline
Philip Massinger & The City Madam, \textbf{\textit{The Duke of Milan}}, The Maid of Honour, The Bondman, The Unnatural Combat, The Renegado or The Gentleman of Venice, A New Way to Pay Old Debts, The Roman Actor, The Great Duke of Florence, The Picture, The Emperor of the East, The Guardian, The Bashful Lover \\
\hline
Thomas May & \textbf{\textit{The Heir}}, Cleopatra --- Queen of Egypt, Julia Agrippina --- Empress of Rome \\
\hline
Thomas Middleton & The Phoenix, Michaelmas Term, A Trick to Catch the Old One, A Mad World My Masters, The Puritain or The Widow of Watling Street, \textbf{\textit{Your Five Gallants}}, The Widow, The Mayor of Quinborough, A Chaste Maid in Cheapside, More Dissemblers Beside Women, Women Beware Women, A Game at Chess \\
\hline
Thomas Nabbes & \textbf{\textit{Covent Garden}}, Tottenham Court, Hannibal and Scipio, The Bride, The Unfortunate Mother \\
\hline
\makecell[l]{William\\Shakespeare} & The Comedy of Errors, Richard III, The Taming of the Shrew, The Two Gentlemen of Verona, Romeo and Juliet, Richard II, King John, The Merchant of Venice, Henry IV --- Part 1, Henry IV --- Part 2, Much Ado About Nothing, Henry V, Julius Caesar, As You Like It, Twelfth Night, Hamlet, Merry Wives of Windsor, Troilus and Cressida, Othello, Measure for Measure, Macbeth, King Lear, \textbf{\textit{Antony and Cleopatra}}, Coriolanus, Cymbeline, The Tempest \\
\hline
James Shirley & The School of Compliment, The Maid's Revenge, The Wedding, The Witty Fair One, The Grateful Servant, The Humorous Courtier, Love's Cruelty, The Ball, The Traitor, Hyde Park, Changes or Love in a Maze, The Bird in a Cage (The Beauties), The Young Admiral, The Gamester, The Opportunity, The Example, The Lady of Pleasure, The Coronation, The Duke's Mistress, The Royal Master, The Doubtful Heir, The Constant Maid, The Gentleman of Venice, Saint Patrick for Ireland --- Part 1, The Politician, The Arcadia, The Imposter, \textbf{\textit{The Sisters}}, The Cardinal, The Brothers, The Court Secret \\
\hline
John Webster & The White Devil (Vittoria Corombona), \textbf{\textit{The Duchess of Malfi}}, The Devil's Law Case (When Women Go to Law the Devil is Full of Business) \\
\hline
Robert Wilson & \textbf{\textit{The Three Ladies of London}}, The Three Ladies of London, The Cobbler's Prophecy \\
\hline
\end{longtable} 

\section{Disputed and Co-Authored Corpus Contents}

All plays in the disputed and co-authored corpus by the author they were attributed to in the original corpora.

\begin{longtable}{|p{0.2\linewidth} | p{0.75\linewidth}|}
\hline
\textbf{Labeled Author} & \textbf{Plays} \\
\hline
George Chapman & Sir Giles Goosecap, Two Wise Men and All the Rest Fools \\
\hline
Thomas Dekker & Patient Grissel, The Wonder of a Kingdom \\
\hline
John Ford & The Laws of Candy, The Queen \\
\hline
Henry Glapthorne & Revenge for Honor (The Parricide) \\
\hline
Robert Greene & George a Green the Pinner of Wakefield \\
\hline
Thomas Heywood & The Fair Maid of the Exchange \\
\hline
John Marston & Histriomastix or The Player Whipped, The Insatiate Countess \\
\hline
Thomas Middleton & Anything for a Quiet Life, The Family of Love \\
\hline
\makecell[l]{William\\Shakespeare} & Henry VI --- Part 1, Henry VI --- Part 2, Henry VI --- Part 3, Henry VIII, Pericles --- Prince of Tyre, Timon of Athens, Titus Andronicus, The Two Noble Kinsmen \\
\hline
John Webster & Appius and Virginia, The Thracian Wonder \\
\hline
\end{longtable}

\section{Comparison Corpus Contents}

All plays in the comparative corpus by author. Plays that were attributed to Shakespeare by the model are bolded and italicized.

\begin{longtable}{|p{0.2\linewidth} | p{0.75\linewidth}|}
\hline
\textbf{Author} & \textbf{Plays} \\
\hline
Robert Armin & The Two Maids of More-Clacke \\
\hline
Thomas Baker & The Fine Lady's Airs \\
\hline
\makecell[l]{James\\Nelson Barker} & \textbf{\textit{The Indian Princess}} \\
\hline
J. M. Barrie & \textbf{\textit{Dear Brutus}}, \textbf{\textit{Peter Pan}} \\
\hline
Lording Barry & Ram Alley \\
\hline
Barnabe Barnes & The Devil's Charter \\
\hline
Clifford Bax & \textbf{\textit{Square Pegs}} \\
\hline
Francis Beaumont & The Knight of the Burning Pestle \\
\hline
Dabridgecourt Belchier & Hans Beer-Pot (See Me and See Me Not) \\
\hline
Arnold Bennett & \textbf{\textit{The Great Adventure}} \\
\hline
William Berkeley & The Lost Lady \\
\hline
\makecell[l]{Hugh Henry\\Brackenridge} & The Battle of Bunkers Hill \\
\hline
Alexander Brome & The Cunning Lovers \\
\hline 
Robert Browning & \textbf{\textit{A Blot in the Scutcheon}} \\
\hline
Henry Burnell & Landgartha \\
\hline
Lodowick Carlell & The Deserving Favorite \\
\hline
\makecell[l]{Richard\\Claude Carton} & \textbf{\textit{Lady Huntworth's Experiment}} \\
\hline
William Cartwright & The Royal Slave \\
\hline
William Cavendish & The Country Captain, The Variety \\
\hline
Susanna Centlivre & The Busie Body, The Perjur'd Husband \\
\hline
\makecell[l]{Robert\\Chamberlain} & The Swaggering Damsel \\
\hline
George Coleman & \textbf{\textit{John Bull}} \\
\hline
Abraham Cowley & Love's Riddle \\
\hline
Aleister Crowley & \textbf{\textit{Household Gods}} \\
\hline
Robert Daborne & A Christian Turned Turk \\
\hline
John Denham & The Sophy \\
\hline
Thomas Drue & The Duchess of Suffolk \\
\hline
William Dunlap & \textbf{\textit{Andre}} \\
\hline
Lord Dusany & \textbf{\textit{If}} \\
\hline
Nathan Field & Amends for Ladies, A Woman is a Weathercock\\
\hline
Jasper Fisher & Fuimus Troes (The True Trojans) \\
\hline
Phineas Fletcher & Sicelides \\
\hline
Ralph Freeman & Imperiale \\
\hline
John Galsworthy & \textbf{\textit{A Bit O' Love}}, \textbf{\textit{The Eldest Son}}, \textbf{\textit{A Family Man}}, \textbf{\textit{The First and the Last}}, \textbf{\textit{The Foundations}}, \textbf{\textit{The Fugitive}}, \textbf{\textit{Joy}}, \textbf{\textit{Justice}}, \textbf{\textit{The Little Dream}}, \textbf{\textit{The Little Man}}, \textbf{\textit{Loyalties}}, \textbf{\textit{The Mob}}, \textbf{\textit{The Skin Game}}, \textbf{\textit{Strife}} \\
\hline
Thomas Godfrey & The Prince of Parthia \\
\hline
Johann Wolfgang von Goethe & \textbf{\textit{Faust}} \\
\hline
John Gough & The Strange Discovery \\
\hline
Fulke Greville & Alaham \\
\hline
John Johns & Adrasta \\
\hline
William Kemp & A Knack to Know a Knave \\
\hline
Henry Killigrew & The Conspiracy \\
\hline
John Kirke & The Seven Champions of Christendom \\
\hline
\makecell[l]{James\\Sheridan Knowles} & \textbf{\textit{The Love Chase}} \\
\hline
Thomas Kyd & Soliman and Perseda, The Spanish Tragedy (Hieronimo is Mad Again) \\
\hline
Maurice Kyffin & Andria \\
\hline
William Habington & The Queen of Aragon \\
\hline
Samuel Harding & Sicily and Naples \\
\hline
Joseph Harris & The City Bride \\
\hline
William Haughton & Englishmen for My Money \\
\hline
Peter Hausted & The Rival Friends \\
\hline
William Hawkins & Apollo Shroving \\
\hline
\makecell[l]{Gorges\\Edmond Howard} & \textbf{\textit{The Female Famester}} \\
\hline
Henrik Ibsen & \textbf{\textit{A Doll's House}}, Hedda Gabler \\
\hline
Elizabeth Inchbald & \textbf{\textit{Such Things Are}}, \textbf{\textit{The Widow's Vow}} \\
\hline
Jerome K. Jerome & \textbf{\textit{Fanny and the Servant Problem}}, \textbf{\textit{Woodbarrow Farm}} \\
\hline
\makecell[l]{Henry\\Arthur Jones} & \textbf{\textit{Dolly Reforming Herself}}, \textbf{\textit{Michael and His Lost Angel}} \\
\hline
D. H. Lawrencee & \textbf{\textit{Touch and Go}} \\
\hline
John Leacock & \textbf{\textit{The Fall of British Tyranny}} \\
\hline
Thomas Lodge & The Wounds of Civil War \\
\hline
Samuel Low & \textbf{\textit{The Politician Out-Witted}} \\
\hline
Sir William Lower & The Phoenix in Her Flames \\
\hline
Thomas Lupton & All for Money \\
\hline
James Mabbe & The Spanish Bawd (Calisto and Meliboea) \\
\hline
Charles Macklin & The Covent Garden Theatre \\
\hline
Gervase Markham & The Dumb Knight, Herod and Antipater \\
\hline
Shakerley Marmion & A Fine Companion, Holland's Leaguer \\
\hline
John Mason & The Turk \\
\hline
Jasper Mayne & The City Match \\
\hline
Edward Moore & \textbf{\textit{The Gamester}} \\
\hline
Thomas Morton & \textbf{\textit{Speed the Plough}} \\
\hline
Arthur Murphy & \textbf{\textit{The Grecian Daughter}} \\
\hline
Thomas Newman & The Andrian Woman (Andria), The Eunuch \\
\hline
Mordecai Manuel Noah & \textbf{\textit{She Would Be a Soldier}} \\
\hline
John O'Keeffe & \textbf{\textit{Wild Oats}} \\
\hline
Henry Nevil Payne & The Fatal Jealousie \\
\hline
Arthur Pinero & \textbf{\textit{The Big Drum}}, The 'Mind the Paint' Girl \\
\hline
Henry Porter & The Two Angry Women of Abingdon \\
\hline
Thomas Randolph & \textbf{\textit{The Jealous Lovers}} \\
\hline
Thomas Rawlins & The Rebellion \\
\hline
Nathaniel Richards & Messalina --- The Roman Empress \\
\hline
Robert Rogers & Ponteach: The Savages of America \\
\hline
Edmond Rostand & \textbf{\textit{Cyrano de Bergerac}} \\
\hline
Samuel Rowley & The Noble Spanish Soldier (The Noble Soldier or A Contract Broken Justly Revenged), When You See Me You Know Me (Henry the Eighth) \\
\hline
Joseph Rutter & The Shepherds' Holiday \\
\hline
S. S. & The Honest Lawyer \\
\hline
W. S. & Thomas Lord Cromwell \\
\hline
Edward Sharpham & Cupid's Whirligig, The Fleer \\
\hline
\makecell[l]{George\\Bernard Shaw} & \textbf{\textit{Arms}}, \textbf{\textit{The Devil's Disciple}}, \textbf{\textit{Fanny's First Play}}, \textbf{\textit{Man and Superman}} \\
\hline
John Stephens & Cynthia's Revenge \\
\hline
William Stevenson & Gammer Gurton's Needle \\
\hline
Algernon Charles Swinburne & \textbf{\textit{The Duke of Gandia}}, \textbf{\textit{Erechtheus}}, \textbf{\textit{Rosamund}} \\
\hline
Robert Tailor & The Hog Hath Lost His Pearl \\
\hline
Brandon Thomas & \textbf{\textit{Charley's Aunt}} \\
\hline
Thomas Tomkis & Albumazar, Lingua or The Combat of the Tongue and the Five Senses of Superiority \\
\hline
Cyril Tourneur & The Atheist's Tragedy \\
\hline
Royall Tyler & \textbf{\textit{The Contrast}} \\
\hline
Nicolas Udall & Ralph Roister Doister \\
\hline
George Wapull & The Tide Tarrieth No Man \\
\hline
Oscar Wilde & \textbf{\textit{Vera}}, \textbf{\textit{A Woman of No Importance}} \\
\hline
George Wilkins & The Miseries of Enforced Marriage \\
\hline
Nathaniel Woodes & The Conflict of Conscience \\
\hline
Robert Yarington & Two Lamentable Tragedies \\
\hline
Richard Zouch & The Sophister \\
\hline
\end{longtable}

\end{document}